\documentclass[10pt, journal, compsoc]{IEEEtran}
\usepackage[nocompress]{cite}
\usepackage{amssymb}
\usepackage[backref]{hyperref}
\usepackage{graphicx}
\usepackage{bbding}
\usepackage{booktabs}
\usepackage{pifont}
\usepackage{amsmath}
\usepackage{soul}
\usepackage{color, xcolor}

\def\etal{\emph{et al}.}

\def\ie{\emph{i.e}.}

\def\hl{}
\def\hll{}

\soulregister{\cite}7
\soulregister{\ref}7

\begin{document}
\title{\hl{HDhuman: High-quality Human Novel-view Rendering from Sparse Views}}

\author{Tiansong Zhou$^{\dagger}$, Jing Huang$^{\dagger}$, Tao Yu, Ruizhi Shao, Kun Li$^{*}$

\IEEEcompsocitemizethanks{
\IEEEcompsocthanksitem $\dagger$ Equal contribution.
\IEEEcompsocthanksitem $*$ Corresponding author.
\IEEEcompsocthanksitem Kun Li, Tiansong Zhou, and Jing Huang are with Tianjin University. Email: lik@tju.edu.cn, tiansong97@tju.edu.cn, hj00@tju.edu.cn.
\IEEEcompsocthanksitem Tao Yu and Ruizhi Shao are with Tsinghua University. Email: ytrock@126.com, shaorz20@mails.tsinghua.edu.cn.
\IEEEcompsocthanksitem This work was done when Tiansong Zhou was an intern at Tsinghua University.}
}

\IEEEtitleabstractindextext{%
\begin{abstract}
\hl{In this paper, we aim to address the challenge of novel view rendering of human performers who wear clothes with complex texture patterns using a sparse set of camera views. 
Although some recent works have achieved remarkable rendering quality on humans with relatively uniform textures using sparse views, the rendering quality remains limited when dealing with complex texture patterns as they are unable to recover the high-frequency geometry details that are observed in the input views.
To this end, we propose HDhuman, which uses a human reconstruction network with a pixel-aligned spatial transformer and a rendering network with geometry-guided pixel-wise feature integration to achieve high-quality human reconstruction and rendering. The designed pixel-aligned spatial transformer calculates the correlations between the input views and generates human reconstruction results with high-frequency details. Based on the surface reconstruction results, the geometry-guided pixel-wise visibility reasoning provides guidance for multi-view feature integration, enabling the rendering network to render high-quality images at 2k resolution on novel views. Unlike previous neural rendering works that always need to train or fine-tune an independent network for a different scene, our method is a general framework that is able to generalize to novel subjects. Experiments show that our approach outperforms all the prior generic or specific methods on both synthetic data and real-world data. Source code and test data will be made publicly available for research purposes at {http://cic.tju.edu.cn/faculty/likun/projects/HDhuman/index.html.}}
\end{abstract}

}

\maketitle

\section{Introduction}
\hl{Realistic free-viewpoint rendering of human performers is in increasing demand with the development of AR/VR. Neural radiance fields (NeRF) \cite{nerf} is the most promising way that render photo-realistic images on novel views. However, NeRF needs a lot of input views to produce photo-realistic novel-view images. When the input views are highly sparse, the rendering quality of NeRF will degrade dramatically. }

\hl{To render novel-view images of human performers from sparse views, neural body \cite{neuralbody} integrates SMPL model \cite{smpl} to the NeRF framework, in which they anchor a latent code on each vertex of SMPL model to integrate the observations over video frames.}
\hl{However, as the SMPL model does not contain any geometry details such as clothing folds, neural body is difficult to capture high-frequency details of human performers, making its rendering quality remain limited when dealing with human performers that wear loose clothes with complex texture patterns.}
Moreover, neural body\cite{neuralbody} needs to train an independent network for each human and the training procedure is extremely time-consuming (at least 10 hours for each subject), which further limits its applications.

\begin{figure}[t]
    \centering
    \includegraphics[width=\linewidth]{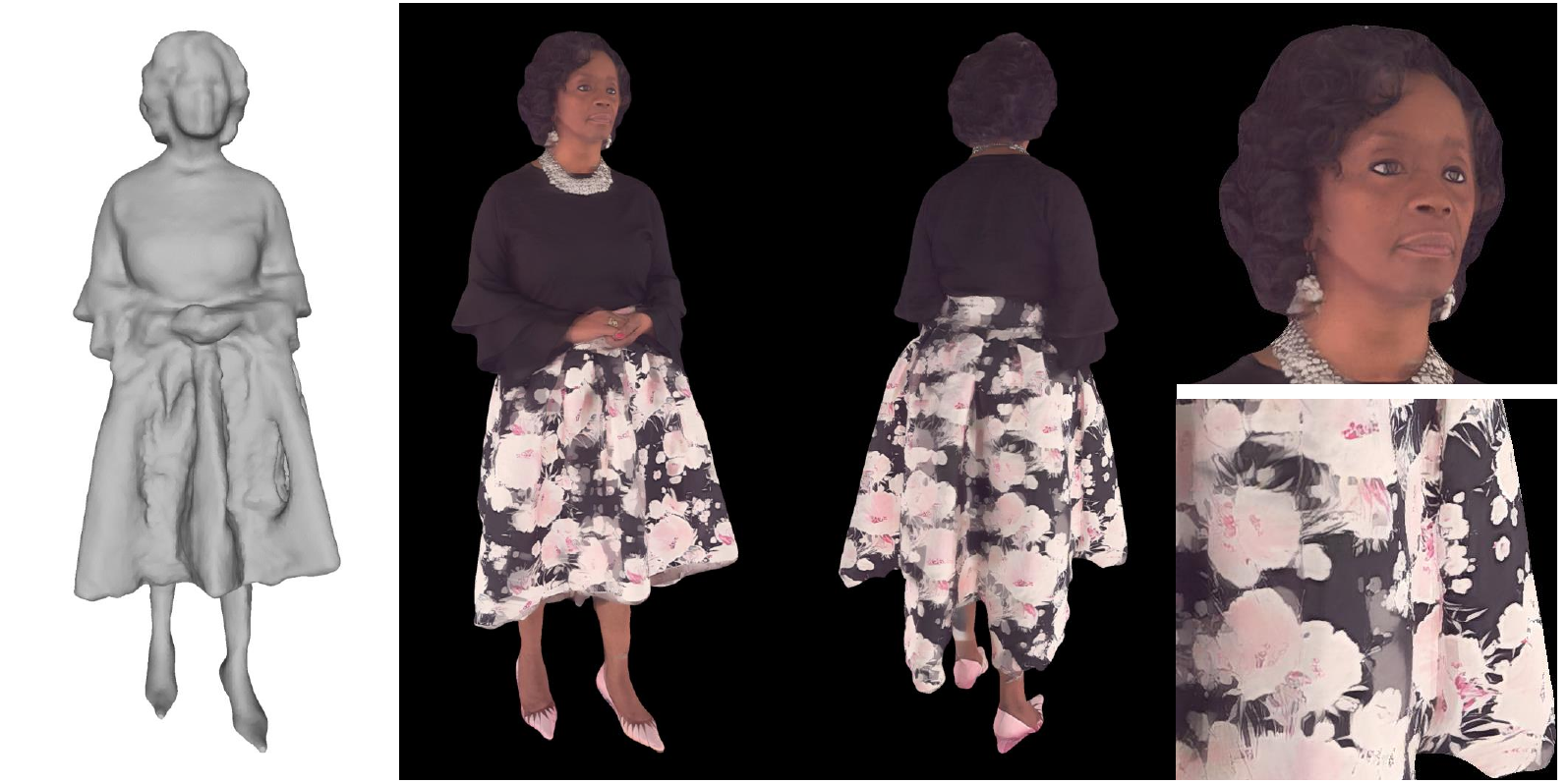}
    \caption{High-quality reconstruction and rendering of a challenging human performer that wears a long dress with complex texture patterns from only 6 input views, without any fine-tuning.}
    \label{fig:teaser}
\end{figure}

In this work, we propose HDhuman, a general method that is able to render high-quality human images at 2k resolution using sparse views. The used camera views are no more than 8 views in a uniformly distributed circle arrangement. 
As shown in Fig. \ref{fig:teaser}, we are able to handle challenging human subjects wearing long dresses with complex texture patterns.
\hl{HDhuman uses a reconstruction-rendering pipeline to achieve the goal of high-quality novel-view rendering from sparse views, demonstrating that recovering geometry details observed in the input views significantly helps to improve the rendering quality.} Firstly, We design a reconstruction network to reconstruct highly-detailed human models, recovering the geometry details of human performers, such as clothing folds. Then, based on the reconstruction results, we propose a rendering network to render high-quality images on novel views at 2k resolution.

\hl{For human reconstruction from sparse views, the key is to fuse the geometry details that are observed in the input multi-view images. Instead of using a naive averaging pooling operation for multi-view feature fusion in PIFu\cite{pifu}, we present a pixel-aligned spatial transformer in the reconstruction network. The proposed transformer is able to automatically calculate different fusion weights for each input view. In this way, we are able to preserve the geometry details such as clothing folds that are observed in the input images, resulting in highly-detailed reconstruction.}

\hl{To achieve the goal of high-quality rendering from sparse views, the core is to solve the severe occlusion problems caused by the sparsity of input views. If we warp all the pixels of source views to the novel view, the occlusion will cause artifacts and blurring in the rendering results, as many pixels are invisible in the novel view. To solve this problem, we propose geometry-guided pixel-wise feature integration in the rendering network. Based on the reconstruction results from the reconstruction network, the proposed integration method will perform visibility reasoning and find the visible source views for each pixel of the novel view, enabling us to solve the occlusion problems and get high-quality rendering images.}

Unlike the neural-network-based rendering works that need to train or fine-tune an independent network for each scene or human \cite{nerf, neuralbody, dynamicTexture, neuraltexture}, the proposed HDhuman is a general rendering framework. Benefiting from the generalization ability of pixel-aligned spatial transformer and geometry-guided feature integration, our method is able to perform high-quality reconstruction and rendering for novel subjects without any fine-tuning. Experiments show that our general approach significantly improves the rendering quality compared with all the existing generic and specific works. 

\hl{Overall, thanks to the pixel-aligned spatial transformer for highly-detailed human reconstruction and the geometry-guided pixel-wise feature integration for solving the occlusion problems, we are able to achieve the goal of high-quality human novel-view rendering from sparse views.} In summary, our main contributions are:
\begin{itemize}
    \item We propose a new general neural rendering framework capable of rendering novel views of human performers with complex texture patterns at 2k image resolution from a sparse set of camera views.
    \item A pixel-aligned spatial transformer that is able to perform efficient multi-view feature fusion in human reconstruction, enabling us to reconstruct highly-detailed human models from sparse views.
    \item A geometry-guided pixel-wise feature integration method for efficiently solving the severe occlusion problems that are caused by the sparsity of input views.
    \item We demonstrate significant rendering quality improvements of our method compared to prior works, especially for humans with loose clothes or complex texture patterns.
\end{itemize}
\section{Related Work}
\subsection{Human Reconstruction}
Human reconstruction has a long and overlapping history in both computer vision and graphics. In earlier years, multi-view stereo (MVS) is the most widely used method for human reconstruction. Liu \etal \cite{continuousdepth} proposes a continuous depth map estimation method for improving the performances of MVS. 

Over the past decade, benefiting from the development of hardware, many works \cite{highquality, fusion4d, function4d, realtime} take single or multi RGBD views as input to get amazing real-time high-fidelity human reconstruction and rendering. However, the relied depth sensors are only able to capture humans or objects that are near to the sensors. When humans move far from the sensors (farther than 3m), the depth sensors will fail to capture depth streams, which greatly limits their applications. 

\hl{In recent years, human reconstruction from only a single RGB camera \cite{fof, image, learning, pifu, pifuhd, livecap, monoperfcap, smplicit} draws great research attention.} The most promising method for single-view human reconstruction is the pixel-aligned implicit function (PIFu) \cite{pifu}. PIFu takes a single color image as input and encodes the image to generate pixel-aligned image features. Based on the pixel-aligned features that contain the geometry details in the input image, it learns an implicit function over the 3D space. Following PIFu, PIFuHD \cite{pifuhd} introduces a multi-level framework for high-fidelity 3D reconstruction of clothed humans. It uses an additional image translation network to predict normal maps from original color images. The predicted normal maps enable it to reconstruct more geometry details, such as clothing folds. Moreover, a fine level is used in PIFuHD to recover more subtle details from the high-resolution input images. \hl{Video-based methods \cite{livecap, monoperfcap, monoclothcap} reconstruct dense, space-time coherent deforming geometry of dynamic humans from monocular videos, but they need a pre-captured template mesh. }

\begin{figure*}[ht]
    \centering
    \includegraphics[width=0.85\linewidth]{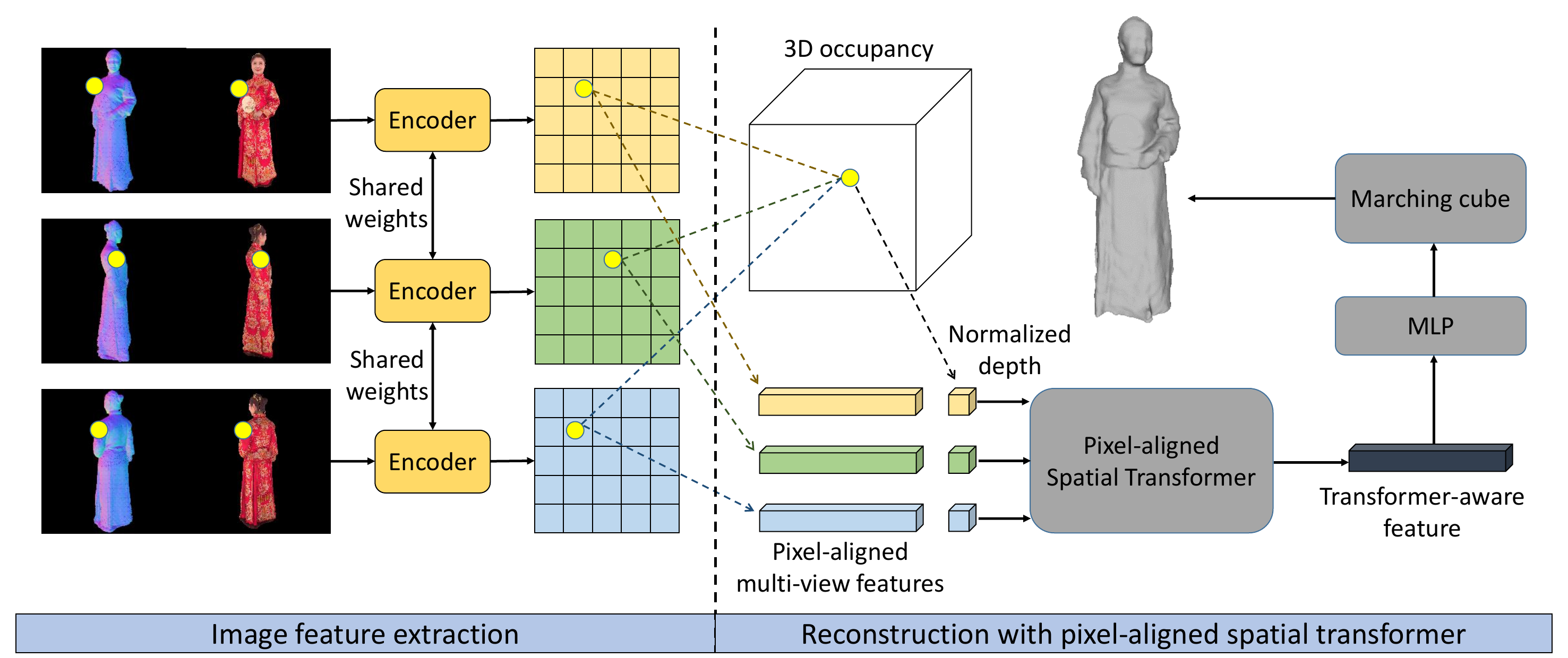}
    \caption{\textbf{Architecture of our reconstruction network.} For memory efficiency, we only use a single level for reconstruction. The pixel-aligned spatial transformer enables us to preserve the geometry details that are observed in the input views, resulting in highly-detailed reconstruction.
    Our framework doesn't use any geometry prior, and hence our reconstruction results are fully pixel-aligned. }
    \label{fig:reconstruction}
\end{figure*}

For multi-view human reconstruction, PIFu uses a naive average pooling operation for multi-view feature fusion, which is not efficient enough to fuse the geometry details that are observed in the multi-view input images. 
\hll{Zins \emph{et al}. \cite{zins} use an attention-based fusion method to automatically weight the input views for multi-view human reconstruction. 
But they use an additional local 3D context encoding layer to make the network aware of the local 3D context around each querying point.
This layer makes the reconstruction more robust but it increases the computation and memory consumption of the networks.
In our observation, we found that directly predicting the SDF value of each point without any ``local information'' is also able to produce high-quality reconstruction results, helping us to strike a good balance between reconstruction quality and computation consumption.
}
Zheng \etal \cite{deepmulticap} use SMPL as the geometry prior to solve the occlusion problem in multi-person reconstruction, but the usage of SMPL not only makes the reconstruction results might be pixel-misaligned, but also makes it unable to reconstruct the human cases with loose clothes, such as long dresses. 
\hll{In contrast, benefiting from the proposed pixel-aligned spatial transformer, we are able to produce highly-detailed reconstruction results without any human template prior such as SMPL, and we are able to handle human performers that wear loose clothes with complex texture patterns.} 

~\\
\subsection{Human Novel-view Rendering}
Many works \cite{deepblending, depthsynthesis, ignor} use the traditional image-based rendering or view synthesis pipeline for novel view rendering, \ie, proxy geometry reconstruction, warping, and blending. To bridge the gap between reconstruction and rendering, Liu \etal \cite{pointmvs} present a point-cloud-based method to improve the multi-view stereo algorithm for more realistic free-view synthesis (FVV). Benefiting from the significant research progress of deep learning in recent years, some works \cite{deepblending, fvs} use neural networks to blend the novel views from the warped source views. However, these warping-based works need dense camera views as input to warp the source views to the novel view. If the input camera views are highly sparse, the performances of this classic method will degrade dramatically as the wide camera baselines will cause severe occlusion problems. 
To tackle this problem, Suo \etal \cite{neuralhumanfvv} proposes a neural blending scheme with neural geometry prior for novel-view rendering of human performers.

More recently, volume rendering is another method to achieve photo-realistic rendering. It uses volumes to represent the shape and appearance of a scene. Multi-plane images (MPIs) \cite{stereomag} is one of the most promising ways for this volumetric representation, which computes a separate RGB$\alpha$ image for each depth plane. The $\alpha$ channel and the RGB channels represent the shape and appearance of the scene respectively. Based on the MPIs-representation, Soft3D \cite{soft3d} introduces a soft 3D representation of the scene geometry. This representation enables it to handle the depth uncertainty of the scene. Benefiting from the significant research progress on neural networks, Flynn \etal and Mildenhall \etal \cite{llff, deepview} use neural networks to predict the MPIs for each input view. 

Another promising research direction for volume rendering is neural radiance fields (NeRF) \cite{nerf}, which uses a multi-layer perceptron (MLP) to represent the density and color fields of a scene and thus it is well-suited for differentiable rendering. \hl{Based on NeRF, Liu \emph{et al}. \cite{nsvf} propose neural sparse voxel structure for faster training and inference.} However, both MPIs-based and NeRF-based \cite{nerf, nsvf} works need dense camera views as input and always need to train or fine-tune an independent network of each scene. 

\hll{
For human rendering from sparse views, some works \cite{StructLocalNerf, HumanNerf_Zhao, SurfaceAlignNerf, AnimatableNerf, EasyMointeracCap} use SMPL template as a prior, which helps to constrain the motion space and improve the rendering quality. 
However, as the SMPL model does not contain any geometry details such as clothing folds, it is difficult for them to capture high-frequency details of human performers, which limits their rendering quality when dealing with human performers with complex texture patterns. 
In contrast, the proposed pixel-aligned spatial transformer in our work enables us to capture the geometry details that are observed in the input views, which helps improve the rendering quality on the areas with complex geometry patterns, such as loose clothes or clothing folds.
Another line of works \cite{HumanNerf_Weng, dnerf} use a deformation field to render human performers from single or multi-view videos, in which they use an MLP to map all the frames of the video to a canonical space and perform free-view rendering using NeRF on the canonical space.
However, as the capacity of a deformation field (MLP) is restricted, the deformation field is only able to handle a limited number of frames. When dealing with a long sequence with complex human motions, the deformation field will be ``overwhelmed'', resulting in rendering failures.
Instead of using a deformation field, we treat each frame as an independent subject to process dynamic cases, enabling us to handle videos with arbitrary lengths. Experiments show that we are able to produce spatial-temporal coherent rendering results naturally when dealing with dynamic human performers.
}

\begin{figure*}[t]
    \centering
    \includegraphics[width=0.85\linewidth]{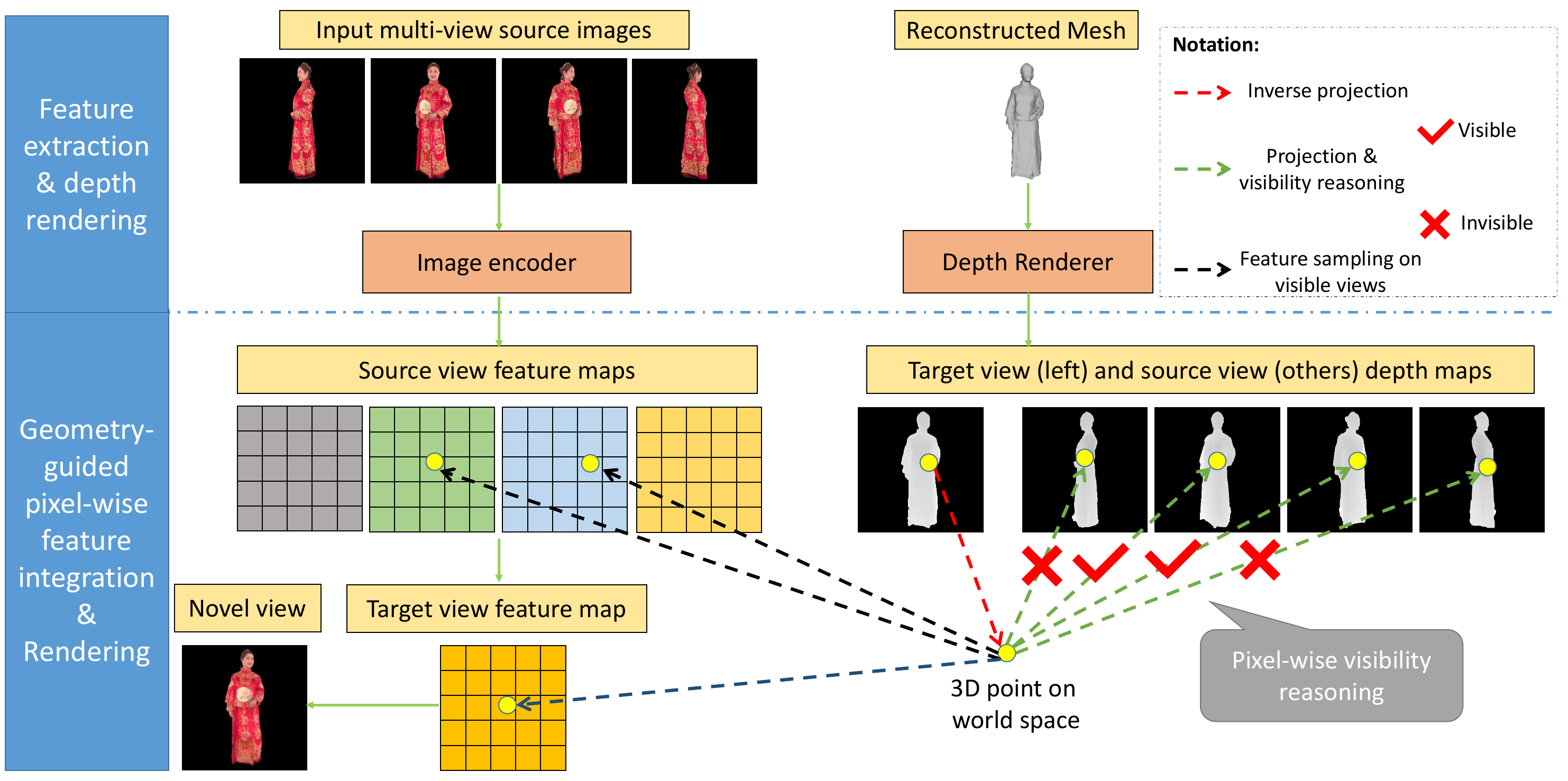}
    \caption{\textbf{Architecture of our rendering network.} The geometry-guided pixel-wise feature integration enables us to solve the severe occlusion problems caused by the sparsity of input views, resulting in high-quality novel view rendering.}
    \label{fig:render}
\end{figure*}

\section{Approach}
To achieve the goal of rendering high-quality human novel view images, we first propose a reconstruction network to reconstruct highly-detailed human models from sparse views (Sec. \ref{sec:reconstruction}). Based on the reconstruction results, we propose our rendering network to render high-quality images at 2k resolution on novel views (Sec. \ref{sec:render}).

\subsection{Human Reconstruction}
\subsubsection{Preliminary}
PIFu\cite{pifu} introduces a pixel-aligned implicit function to reconstruct the underlining 3D human geometry from single or multi-view images. The proposed pixel-aligned implicit function consists of a convolutional image encoder $g$ and a continuous implicit function $f$ that is represented by a multi-layer perceptron. Concretely, the human surface is defined as
\begin{equation}
    f(g(I(x)), z(X)) = s : s \in \mathbb{R},
\end{equation}
where $X$ is the given 3D point, $x = \pi(X)$ is its orthogonal 2D projection, $g(I(x))$ is the bilinear sampled image feature at $x$. During the inference, 3D space is uniformly sampled to infer the occupancy and the final iso-surface is extracted with a threshold of 0.5 using marching cube algorithm \cite{marchingcube} \hl{(where the occupancy value $s$ smaller than 0.5 is inside and $s$ greater than 0.5 is outside)}.

To achieve high-fidelity 3D reconstruction of a clothed human from a single image, recovering detailed information such as clothing folds, PIFuHD\cite{pifuhd} proposes a coarse-to-fine framework for 3D clothed human reconstruction using images with resolution of 1024 $\times$ 1024. It uses an additional network to predict normal maps from the input color images. Compared with PIFu, the predicted normal maps enable it to recover more details as the geometry information in normal maps is more explicit. Moreover, a fine level is used in PIFuHD to recover details from high-resolution images. Specifically, the fine level can be denoted as
\begin{equation}
    f^H(g^H(I_H, F_H, B_H, x_H), \Omega(X)) = s : s \in \mathbb{R},
\end{equation}
where $I_H, F_H, B_H$ are the input color images and the predicted front and back normal maps at the resolution of 1024$\times$1024. $\Omega(X)$ is the feature extracted from the coarse level.

For multi-view reconstruction, PIFu\cite{pifu} uses a naive average pooling operation to fuse multi-view features. 

The average pooling operation in PIFu is not efficient enough to aggregate geometry details that are observed in the input images, as it regards each view equally, resulting in losing subtle geometry details and limiting the rendering quality. Therefore, we need to find a way that is able to calculate the correlations between the input views, providing high-level information for feature aggregation and enabling us to preserve the subtle details.

\begin{table}
    \caption{\textbf{Quantitative reconstruction results on Twindom dataset.} For each model, we use 6 views for reconstruction. Our reconstruction results outperform neural body by a large margin.}
    \centering
    \small
    \scalebox{0.85}{\begin{tabular}{c|cc|cc}
    \toprule
    & \multicolumn{2}{c}{P2S} & \multicolumn{2}{|c}{Chamfer} \\
    & Ours & Neural body\cite{neuralbody} & Ours & Neural body\cite{neuralbody} \\
    \hline
    Model1 & \textbf{0.1866} & 0.9299 & \textbf{0.1968} & 0.7030 \\
    Model2 & \textbf{0.2604} & 0.8902 & \textbf{0.3817} & 0.6753 \\
    Model3 & \textbf{0.1718} & 0.6639 & \textbf{0.1840} & 0.5410 \\
    Model4 & \textbf{0.3858} & 1.8421 & \textbf{0.4070} & 1.5791 \\
    Model5 & \textbf{0.1660} & 0.6104 & \textbf{0.1794} & 0.5164 \\
    \hline
    Average & \textbf{0.2698} & 0.8030 & \textbf{0.2341} & 0.9873 \\
    \bottomrule
    \end{tabular}}
    
    \label{table:recon_twindom}
\end{table}

\begin{table*}[t]
    \caption{\hll{\textbf{Quantitative rendering results on Twindom dataset.} "NB" means neural body. "D+S3D" means Soft3D with our depth maps as input. "NT" means neural texture. "SA-NeRF" means Surface-Aligned NeRF. For each model, we use 6 views as input and 30 views for evaluation. $\uparrow$ means higher is better and $\downarrow$ means lower is better.}}
    
\centering
\small
    \scalebox{0.85}{
        \begin{tabular}{c|ccccc|ccccc|ccccc}
        \toprule
        & \multicolumn{5}{c |}{LPIPS $\downarrow$} & 
        \multicolumn{5}{c |}{SSIM $\uparrow$} & 
        \multicolumn{5}{c }{PSNR $\uparrow$ }\\
        & Ours & NB & D+S3D & NT & SA-NeRF &
        Ours & NB & D+S3D & NT & SA-NeRF & Ours & NB & D+S3D & NT & SA-NeRF \\
        \hline
        Model1 & 
        \textbf{0.146} & 0.266 & 0.180 & 0.295 & 0.294 &
        \textbf{0.876} & 0.842 & 0.831 & 0.790 & 0.795 &
        \textbf{25.32} & 24.63 & 23.84 & 22.96 & 20.62 \\
        Model2 & 
        \textbf{0.196} & 0.290 & 0.252 & 0.391 & 0.360 &
        \textbf{0.782} & 0.731 & 0.701 & 0.612 & 0.633 &
        17.53 & \textbf{17.81} & 16.26 & 11.23 & 15.89 \\
        Model3 & 
        \textbf{0.122} & 0.213 & 0.179 & 0.320 & 0.289 & 
        \textbf{0.865} & 0.832 & 0.793 & 0.712 & 0.732 &
        21.75 & \textbf{21.81} & 19.49 & 13.37 & 13.23 \\
        Model4 &
        \textbf{0.230} & 0.425 & 0.265 & 0.438 & 0.475 &
        \textbf{0.610} & 0.469 & 0.560 & 0.501 & 0.417 &
        \textbf{18.53} & 14.35 & 17.52 & 15.16 & 13.23 \\
        Model5 & 
        \textbf{0.130} & 0.274 & 0.172 & 0.304 & 0.320 &
        \textbf{0.794} & 0.745 & 0.739 & 0.696 & 0.684 &
        \textbf{21.76} & 21.33 & 20.30 & 17.66 & 18.21 \\
        \hline                                      
        Average & 
        \textbf{0.165} & 0.293 & 0.210 & 0.349 & 0.348 &
        \textbf{0.785} & 0.724 & 0.725 & 0.662 & 0.652 &
        \textbf{20.98} & 20.03 & 19.48 & 16.08 & 16.48 \\
        \bottomrule
        \end{tabular}
    }
\label{table:twindom_results}
\end{table*}

\subsubsection{Pixel-aligned Spatial Transformer}\label{sec:reconstruction}

To solve the above limitations of the prior works, we propose a pixel-aligned spatial transformer in the reconstruction network. See Fig. \ref{fig:reconstruction} for an illustration.

\begin{figure*}[ht]
    \centering
		\includegraphics[width=\linewidth]{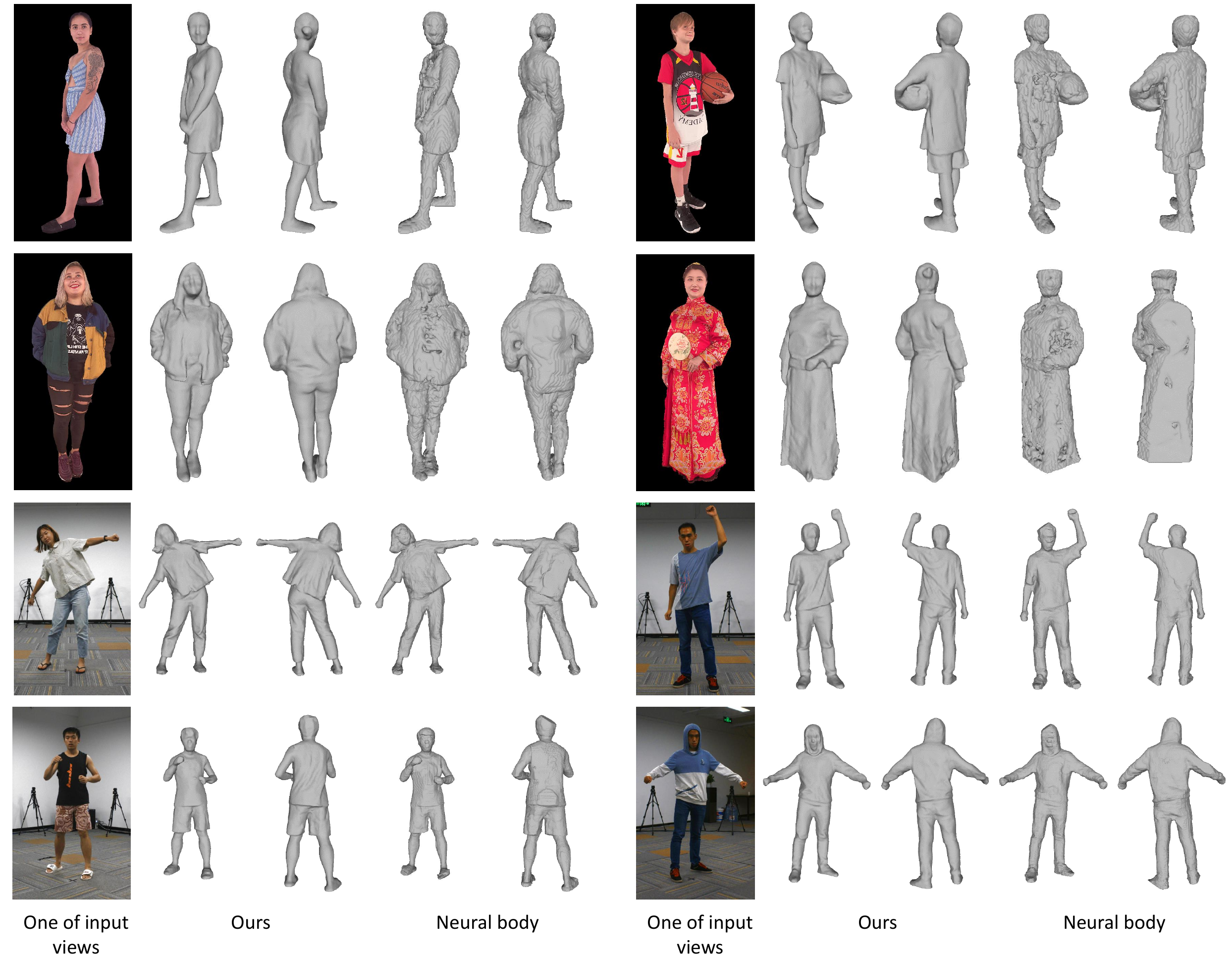}
    \caption{\textbf{Qualitative results of reconstruction.} The top two rows are the results on the Twindom dataset, with 6 views as input. The bottom two rows are the results on real-world data, with 8 views as input. We strongly recommend that readers zoom in on the pages for a better visualization of the reconstructed details. }
	\label{fig:recon_all}
\end{figure*}

\begin{figure*}[ht]
    \centering
		\includegraphics[width=\linewidth]{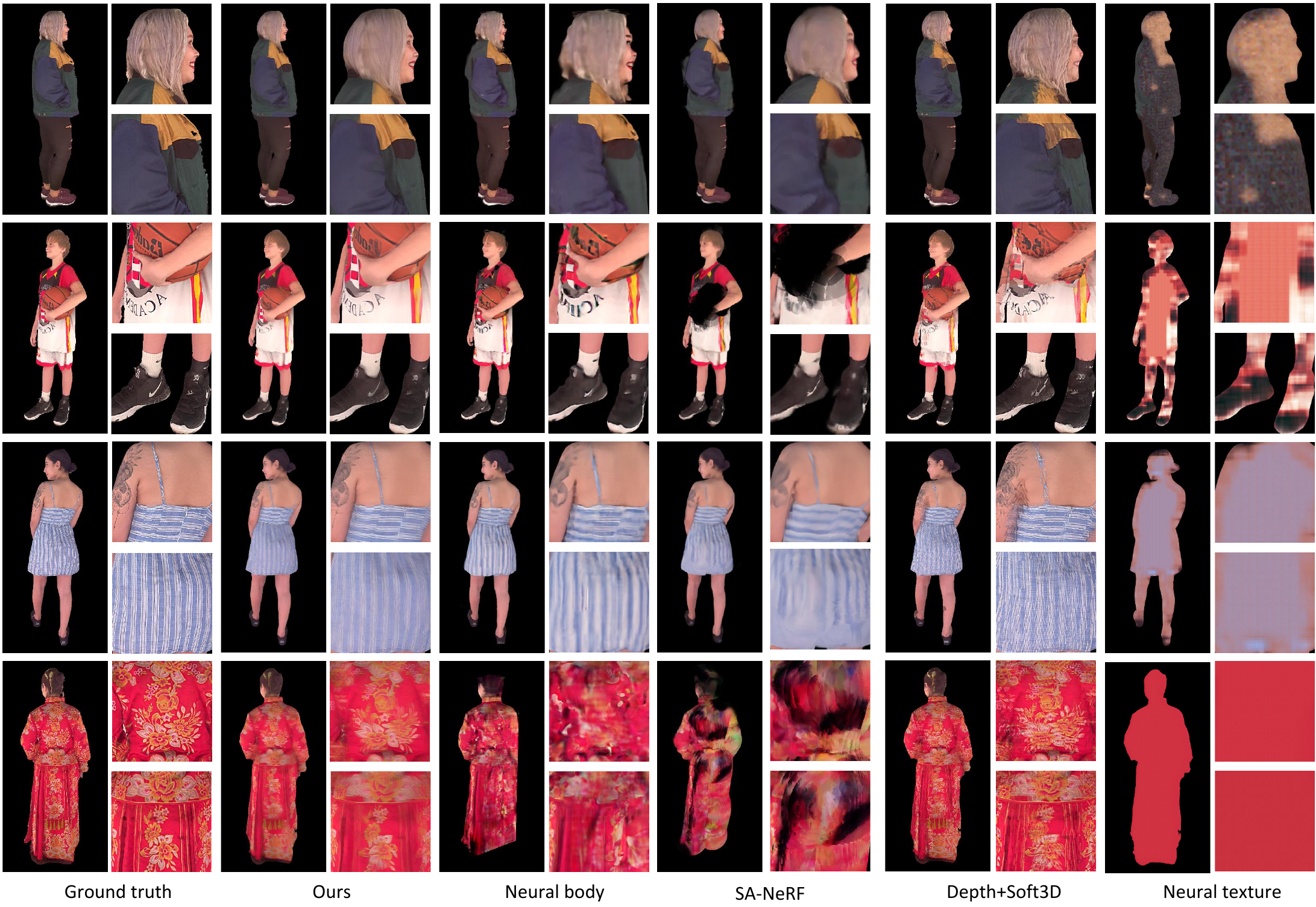}
  \vspace{-0.5cm}
    \caption{\hll{\textbf{Novel view rendering on Twindom dataset.} We use 6 views as input. Results show that our method significantly outperforms all other methods qualitatively, especially in the areas that have complex details.}}
	\label{fig:twindom_results}
\end{figure*}

\begin{figure*}[t]
    \centering
		\includegraphics[width=\linewidth]{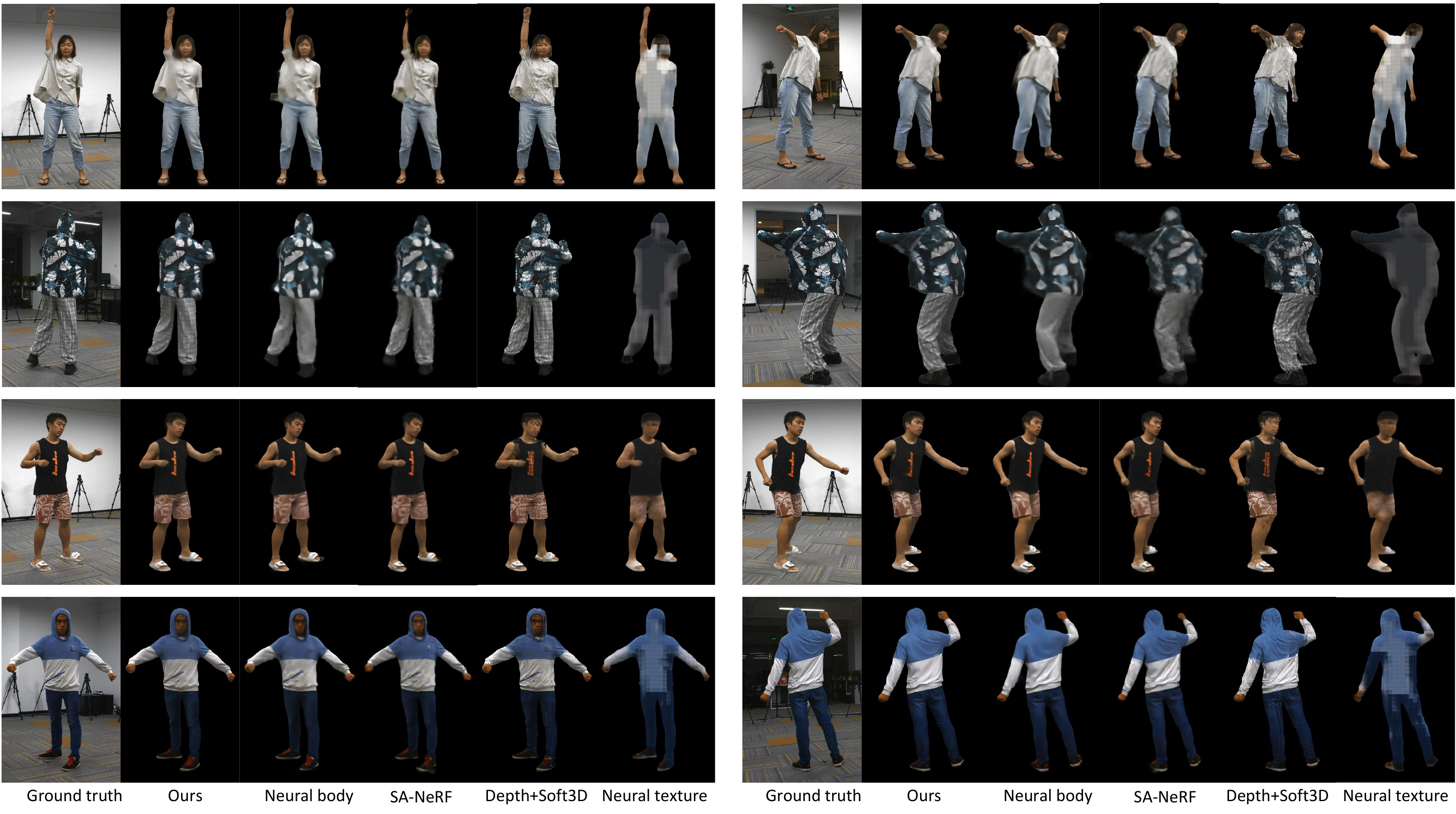}
    \caption{\hll{\textbf{Novel view rendering on our captured multi-view videos dataset.} We use 8 views as input. We strongly recommend the readers to zoom in the pages for a better visualization of the rendered details.}}
	\label{fig:hangzhou_results}
\end{figure*}

\begin{table*}[t]
    \caption{\hll{\textbf{Quantitative rendering results on our multi-view video dataset.} 
    For each frame, we use 8 views as input and 16 views for evaluation. Each sequence has a length of 150 frames.}}
\centering
\small
    \scalebox{0.85}{
        \begin{tabular}{c|ccccc|ccccc|ccccc}
        \toprule
        & \multicolumn{5}{c|}{LPIPS $\downarrow$} & \multicolumn{5}{c|}{SSIM $\uparrow$} & \multicolumn{5}{c}{PSNR $\uparrow$} \\
        & Ours & NB & D+S3D & NT & SA-NeRF &  
        Ours & NB & D+S3D & NT & SA-NeRF & 
        Ours & NB & D+S3D & NT & SA-NeRF \\
        \hline
        Sequence1 & 
        \textbf{0.122} & 0.278 & 0.177 & 0.267 & 0.281 &
        \textbf{0.823} & 0.732 & 0.722 & 0.687 & 0.720 & 
        \textbf{24.14} & 21.05 & 21.11 & 17.61 & 19.50 \\
        Sequence2 &  
        \textbf{0.162} & 0.350 & 0.206 & 0.329 & 0.354 &
        \textbf{0.787} & 0.659 & 0.671 & 0.649 & 0.650 & 
        \textbf{22.36} & 20.60 & 19.13 & 17.11 & 19.51 \\
        Sequence3 & 
        \textbf{0.119} & 0.236 & 0.1723& 0.234 & 0.243 &
        \textbf{0.887} & 0.831 & 0.811 & 0.790 & 0.815 &
        \textbf{25.04} & 22.76 & 22.10 & 19.81 & 21.14 \\
        Sequence4 & 
        \textbf{0.129} & 0.247 & 0.186 & 0.245 & 0.251 &
        \textbf{0.872} & 0.832 & 0.797 & 0.796 & 0.815 & 
        \textbf{26.90} & 25.35 & 24.42 & 22.88 & 23.75 \\
        Sequence5 & 
        \textbf{0.116} & 0.243 &0.174 & 0.259 & 0.246 &
        \textbf{0.880} & 0.822 & 0.800 & 0.751 & 0.809 &
        \textbf{26.47} & 24.32 & 23.45 & 17.59 & 22.11 \\
        \hline                                      
        Average & 
        \textbf{0.129} & 0.271 & 0.183 & 0.267 & 0.275 &
        \textbf{0.850} & 0.775 & 0.760 & 0.735 & 0.761 &
        \textbf{24.98} & 22.82 & 22.04 & 19.00 & 21.20 \\
        \bottomrule
        \end{tabular}
    }
\label{table:hangzhou_results}
\end{table*}

Following PIFuHD \cite{pifuhd}, we first predict frontal normal maps from color images and encode each view's color image and normal map using an hourglass encoder $g$. Our observation is that in multi-view reconstruction, the coarse level network in PIFuHD \cite{pifuhd} is enough to reconstruct high-frequency details with frontal normal maps as input. Therefore, for memory efficiency, we only use a single level for reconstruction \hl{and the input images of the reconstruction network are at a resolution of 512 $\times$ 512}.

To efficiently preserve the geometry details that are observed in the input views, we use a spatial transformer to calculate the correlations between the input multi-view pixel-aligned features. 
Specifically, for a 3D point $X$, given $N$ views encoded features, we stack them together to get $\Phi_{mv}\in\mathbb{R}^{N\times{D}}$, where $D$ is the feature dimensions. Then, we embed it with three learnable different linear layers: 
$\Phi_{q} = \Phi_{mv}W_{q}, \Phi_{k} = \Phi_{mv}W_{k}, \Phi_{v} = \Phi_{mv}W_{v}$, where 
$W_{q}, W_{k}, W_{v}\in\mathbb{R}^{D\times{d_k}}$ 
and $\Phi_{q}, \Phi_{k}, \Phi_{v}\in\mathbb{R}^{N\times{d_k}}$. 
$d_k$ is the embedded feature dimension. After that, a spatial transformer will be applied:
\begin{equation}
\begin{aligned}
    \Phi_{mv\_att}
    &=Transformer(\Phi_{q}, \Phi_{k}, \Phi_{v})\\
    &=softmax(\frac{\Phi_{q}{\Phi_{k}}^{T}}{\sqrt{d_k}}){\Phi_{v}},
    \label{transformer}
\end{aligned}
\end{equation}
where $\Phi_{mv\_att}\in\mathbb{R}^{d_{k}}$ is the transformer-aware multi-view features.
To counter the gradient vanishing problem caused by the softmax operation, we scale the dot-product operation by $\frac{1}{\sqrt{d_k}}$. Different from the feature before transformer calculation, which only contains the geometry details that are observed in a single view, the transformer-aware feature contains all the geometry details that are observed in multi-view images. Compared with the average pooling operation in PIFu \cite{pifu}, the transformer-based fusion operation calculates the correlations between the multi-view features, which provides high-level information for fusion, allowing the network to preserve more geometry details.

To query the depth value of point $X$ with the encoded features, we need to normalize the depth value for each view. Given multi-view human images and the corresponding calibrated camera parameters as input, we will estimate each view's human 2D skeleton using Openpose \cite{openpose}, and then obtain 3D skeleton $S_{w}\in\mathbb{R}^{3\times J}$ in world space through triangulation. $J$ is the number of joints. We use the hip position $H_{w} \in  \mathbb{R}^{3}$ and neck position $N_{w}\in\mathbb{R}^{3}$ to normalize the depth. Specifically,
\begin{equation}
    H^{i}_{c} = R_{i}H_{w} + t_{i} = (H^{i}_{c_{x}}, H^{i}_{c_{y}}, H^{i}_{c_{z}}),
\end{equation}
\begin{equation}
    X^{i}_{c} = R_{i}X + t_{i} = (X^{i}_{c_{x}}, X^{i}_{c_{y}}, X^{i}_{c_{z}}),
\end{equation}
\begin{equation}
    z_{i}(X^{i}_{c_{z}}) = \frac{X^{i}_{c_{z}}-H^{i}_{c_{z}}}
    {\lambda\Vert H_{w} - N_{w} \Vert_2},
    \label{Eq:normalize}
\end{equation}
where $H^{i}_{c}, X^{i}_{c}$ are the hip position and the point $X$ position in view $i$'s camera space respectively. $R_{i} , t_{i}$ are camera $i$'s rotation and translation.  $\lambda$ is a constant value to make sure that the normalized depth always lies in $(0, 1)$, and we set it to $4\sqrt{3}$ in all our experiments.

Finally, we use a multi-layer perceptron to predict the 3D occupancy and use marching cube algorithm to extract the surface from the predicted occupancy, resulting in the final reconstructed mesh.

\subsection{Novel View Rendering} \label{sec:render}
To achieve the goal of high-quality novel view rendering from sparse views, the core problem is to solve the occlusion problems caused by the sparsity of input views. Based on the highly-detailed reconstructed results from the reconstruction network, we propose geometry-guided pixel-wise feature integration to perform efficient visibility reasoning for solving the occlusion problems.  

Concretely, given a sparse set of input images $\{I_{1}, ..., I_{N}\}$, the corresponding calibrated camera parameters and the reconstructed meshes, we use a rendering network to render high-quality novel-view images of human performers. 
The architecture of the rendering network is shown in Fig. \ref{fig:render}. In summary, we use an encode-integration-render framework for novel view rendering, which mainly consists of three parts: 1) encoding each input image to feature space using an encoder-net $e$, 2) pixel-wise visibility reasoning and feature integration using the reconstructed human geometry, and 3) rendering the novel view image using a render-net $r$.

~\\
\textbf{Image encoding.} 
For each input image $I_{n}\in\mathbb{R}^{H\times{W}\times{3}}$, we firstly encode it to high-dimensional feature space and get feature map $F_{n}=e(I_{n})\in\mathbb{R}^{H\times{W}\times{D}}$ at the same resolution. Following Riegler \etal \cite{SVS}, the encoder $e$ uses a Res-UNet architecture, where the encoding part is an ImageNet-pretrained ResNet\cite{he2016deep} and the decoding part upsamples the feature map using nearest-neighbor interpolation, concatenating it with the corresponding feature map (of the same resolution) from the encoding part.

~\\
\textbf{Geometry-guided pixel-wise feature integration.} 
Traditional image-based rendering methods always use warping operation to map the source views to the novel view, in which the warping operation will warp the pixels or areas that are not visible in the novel view, resulting in rendering artifacts or blurring. 
Therefore, we argue that this warping operation is not suitable for solving the occlusion problems and propose our geometry-guided pixel-wise feature integration.
For each pixel in a novel view, the visible input views are different. 
Therefore, we perform visibility reasoning and get the visible source views for each pixel in the novel view. 
Then, we only integrate the features from visible source views to the novel view pixel, the source views that are not visible will be abandoned.

Concretely, given the calibrated camera parameters (intrinsic and extrinsic parameters) of input views $\{C_{1}, ..., C_{N}\}$, the novel view camera $C_{novel}$ and the reconstructed human mesh, we will render depth maps of the input views $\{D_{1}, ..., D_{N}\}$ and novel views $D_{novel}$ using OpenGL or Taichi. Then, for each pixel $p\in\mathbb{R}^{2}$ with valid depth value ($d_{render} > 0$) in the novel view, we will inversely project it to the world space, getting point $P\in\mathbb{R}^{3}$, and then project it to the image space of each input view using the corresponding cameras, getting the re-projected pixel, re-projected depth and rendered depth in each view $\{(p_{reproj_1},d_{reproj_1}, d_{render_1}), ..., (p_{reproj_N}, d_{reproj_N}, d_{render_N})\}$:
\begin{equation}
    (p_{reproj_n}, d_{reproj_n}, d_{render_n}) =  C_{n}(C_{novel}^{-1}(p, d_{render})),
\end{equation}
where $d_{render_n} = D_n(p_{reproj_n})$.
If the difference between the rendered depth and the reprojeted depth is lower than a threshold, we will regard view $n$ to be visible for pixel $p$:
\begin{equation}
| d_{render_n} - d_{{reproj}_n} | < \lambda\cdot min(d_{render_n}, d_{{reproj}_n}),
\end{equation}
where $\lambda$ is a hyper-parameter and we set it to 0.01 in all our experiments.

After the visibility reasoning, we will sample features from all the visible views using bilinear sampling. \ie, $f_n = F_n(p_{reproj_n})$, and getting $\{f_1, ..., f_K\}$, where $K$ is the number of visible source views for pixel $p$. Then, a direction average operation will be performed to integrate source view features into the novel view pixel:
\begin{equation}
    f_{fusion} = \frac{1}{W} \sum_{k=1}^{K}{max(0, \cos\langle dir_{novel}, dir_k \rangle) \cdot f_k},
\end{equation}
where $dir_{novel}, dir_k$ are the novel view direction and the visible source view $k$'s direction respectively. The fused novel view feature is divided by $W = \sum_{k=1}^{K}{max(0, \cos\langle dir_{novel}, dir_{k} \rangle)}$ for normalization.

~\\
\textbf{Novel view rendering.} After getting the feature map $F_{novel}$ of the novel view, we will use a convolutional render-net $r$ to render the novel view's color image \ie, $I_{novel} = r(F_{novel})$. 

\section{Experiments}
\subsection{Training}\label{src:training}
\textbf{Dataset.} We collect 1700 high-quality textured human meshes from Twindom\cite{twindom} 
as a large-scale dataset for the training and evaluation of our reconstruction network and rendering network. The collected models have a wide range of clothing, poses and shapes. We randomly split the models into a training set of 1500 subjects and a testing set of 200 subjects.

~\\
\textbf{Reconstruction network training.} For each subject in the dataset, we generate 360 virtual perspective cameras in the yaw axis (1 camera for each degree) and each camera has a random pitch angle. Then, we render 360 images at a resolution of 512 $\times$ 512 using Taichi. During training, we randomly pick 4 views over the 360 images as the input for each iteration. We use Adam \cite{adam} optimizer with a learning rate of $1 \times {10^{-4}}$ and train the reconstruction network for nearly 300000 iterations for convergence. The training procedure costs nearly 3 days with a batch size of 1 in a single Nvidia TitianXp GPU.

~\\
\textbf{Rendering network training.} The rendering network needs reconstructed meshes as input, but we didn't use the ground-truth geometry. Instead, we randomly pick 5 views over the 360 images and use the reconstruction network to generate a coarse mesh for each subject. With the coarse mesh as input, the rendering network will have greater generalization ability and have more robust performances. We use the generated perspective cameras to render 360 images and depth maps at a higher resolution of $1024 \times 1024$ for each subject in our dataset. We divide the yaw axis degrees into 6 parts, each part has a range of 60 degrees. During training, we randomly pick one view in each part, resulting in 6 images as input. Then, we randomly select one view as the novel view and the other views are the source views. We use Adam optimizer with a learning rate of $1 \times 10^{-4}$ and train for nearly 400000 iterations for convergence. The training procedure costs nearly 2 days with a batch size of 1 using a single Nvidia RTX3090 GPU.

\begin{table}[t]
    \caption{\textbf{Quantitative evaluation for ablation study on Twindom dataset.} }
\centering
\small
    \scalebox{1}
    {
        \begin{tabular}{c|c|c|c}
        \toprule
        & Lpips~$\downarrow$ & SSIM~$\uparrow$
        & PSNR~$\uparrow$ 
        \\
        \hline
        Full model & \textbf{0.1647} & \textbf{0.7853} & 
        \textbf{20.98} \\
        Average pooling & 0.1880 & 0.7610 & 19.82 \\
        No visibility reasoning & 0.1865 & 0.7720 & 20.46 
        \\
        \bottomrule
        \end{tabular}
    
}
\label{table:albation}
\end{table}

\subsection{Evaluation Settings}
In this section, we will introduce our evaluation metrics, compared methods and the settings for each compared method. The evaluation will be performed both on static human performers (synthetic data) and dynamic human performers (real-world data). We highly recommend readers to refer to the supplemental video
for a better visualization of the reconstruction and rendering results. 

~\\
\textbf{Metrics.} For reconstruction, we use Chamfer distance and Point-to-Surface distance (P2S) for quantitative evaluation. For rendering, we quantitatively evaluate our method using the three most widely-used metrics: peak signal-to-noise ratio (PSNR), structural similarity index (SSIM) and learned perceptual image patch similarity (LPIPS).

~\\
\textbf{Compared methods.} As our final goal is human novel-view rendering, we compare our method with both generic and specific rendering methods\cite{neuralbody, soft3d, neuraltexture, SurfaceAlignNerf}: \textbf{1)} Neural texture\cite{neuraltexture} introduces differentiable latent neural texture maps for novel view rendering. It needs to train an independent network for a different subject. Note that, to evaluate neural texture on dynamic humans, topologically consistent reconstruction of human performers is necessary. Therefore, we simply track the reconstructed results and get a topologically consistent mesh sequence for each dynamic human;
\textbf{2)} Soft3D \cite{soft3d} is an MPI-based method for view synthesis, which uses multi-plane images (MPI) to handle the uncertainty of depth maps. It is a generic view synthesis method that doesn't need to train any network. Note that Soft3D doesn't use any geometry prior, and hence it needs to use MPI to estimate an initial depth map for each input image, which is extremely coarse, making the quality of synthetic images very low. For fairness, we use the rendered depth maps in our method as the initial depth maps for an additional input of Soft3D, which improves its performance;
\textbf{3)} Neural body \cite{neuralbody} proposes a NeRF-based implicit neural representation for human novel-view rendering, which needs to train a separate network for each human. Note that reconstruction results are only available on neural body, and hence we will perform comparisons with it for reconstruction evaluation.
\hll{\textbf{4)} Surface-Aligned NeRF \cite{SurfaceAlignNerf} builds neural radiance fields on the surfaces of SMPL meshes, in which it injectively maps a spatial point to a surface-aligned representation that consists of a projected surface point and a signed height to the mesh surface.}


\begin{figure}[t]
    \centering
		\includegraphics[width=\linewidth]{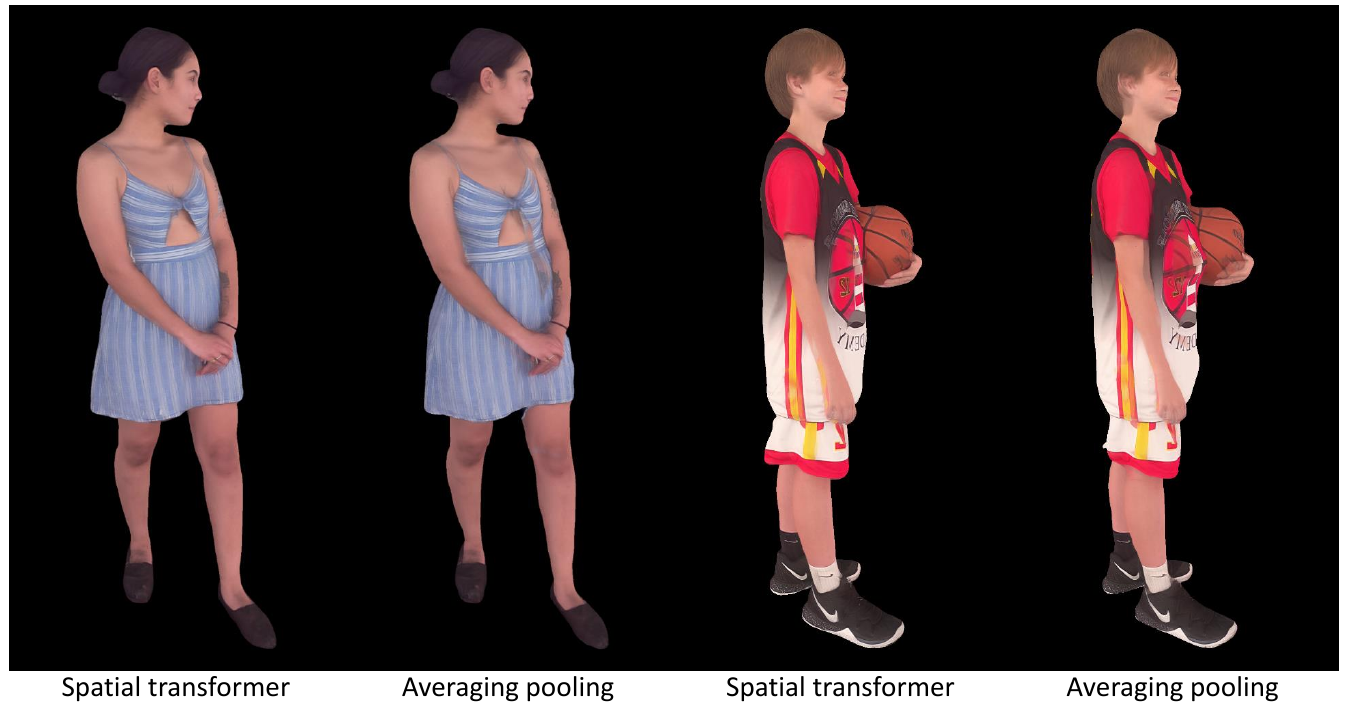}
    \caption{\textbf{Comparisons with average pooling operation.} Compared with the average pooling operation for multi-view feature fusion that was introduced in PIFu, our proposed spatial transformer is able to generate higher-quality results.}
	\label{fig:geometry_ablation}
\end{figure}

\subsection{Results on Synthetic Data}
We select 5 models in the test set of Twindom for the evaluation on synthetic data. For each model, we render 36 images in a circle arrangement at a resolution of 2048 $\times$ 2048. We use 6 uniformly distributed views as input and the remaining 30 views for evaluation.
Table \ref{table:recon_twindom} shows the quantitative reconstruction comparisons of our method with neural body \cite{neuralbody}. For both P2S distance and Chamfer distance, our method outperforms neural body by extremely large margins.
Some qualitative reconstruction results are shown in the top two rows of Fig. \ref{fig:recon_all}. The reconstruction results are produced with only 6 views as input, showing that our method is able to perform highly-detailed reconstruction of challenging humans using only a sparse set of camera views. 


For novel-view rendering, quantitative and qualitative comparisons are illustrated in Table~\ref{table:twindom_results} and Fig.~\ref{fig:twindom_results} respectively, showing that our method is able to render high-quality images of challenging humans with complex patterns or loose clothes, such as long dresses


\subsection{Results on Real-World Data}
To evaluate our method's performance on real-world data, we create a multi-view dataset, which captures 5 dynamic human videos at a resolution of 2660 $\times$ 2300 using a multi-camera system that has 24 calibrated synchronized cameras in a circle arrangement. In contrast to the dataset of neural body \cite{neuralbody}, which only captures humans with relatively uniform textures, we capture human performers with complex texture patterns or loose clothes. We select 8 uniformly distributed cameras as the source input views and use the remaining 16 camera views for testing. All sequences have a length of 150 frames.

Some qualitative results are shown in the bottom two rows of Fig.~\ref{fig:recon_all}. Compared with neural body, our reconstruction results have more high-frequency details, such as clothing folds. Achieving great performances on real-world data means that the proposed pixel-aligned spatial transformer is robust and efficient. 

The quantitative results are shown in Table \ref{table:hangzhou_results}. Our method outperforms all other methods on all the measured metrics. Fig.~\ref{fig:hangzhou_results} illustrates some qualitative results. The novel views rendered by our method contain more texture details compared with other methods. Benefiting from the geometry-guided pixel-wise feature integration, we efficiently solve the occlusion problems caused by the sparsity of input views, resulting in robust high-quality rendering on real-world data.

\subsection{Ablation Studies}

\textbf{Impact of the pixel-aligned spatial transformer.} For comparison, we train a reconstruction network using average pooling operation for multi-view feature fusion. Then, we use the reconstruction results generated by the average pooling operation and our proposed transformer as the input of the rendering network separately. \hl{Note that the reconstruction network with average pooling operation can be regarded as a multi-view PIFuHD method.} Quantitative evaluation results on the Twindom dataset are illustrated in the second and third rows of Table~\ref{table:albation}, showing that the spatial transformer used in our method outperforms average pooling operation on all three metrics. Fig.~\ref{fig:geometry_ablation} shows that benefiting from the inner-product operation, our proposed spatial transformer is able to calculate the correlations between input views, resulting in rendering results with higher quality compared with the average pooling operation.

\begin{figure}
    \centering
		\includegraphics[width=\linewidth]{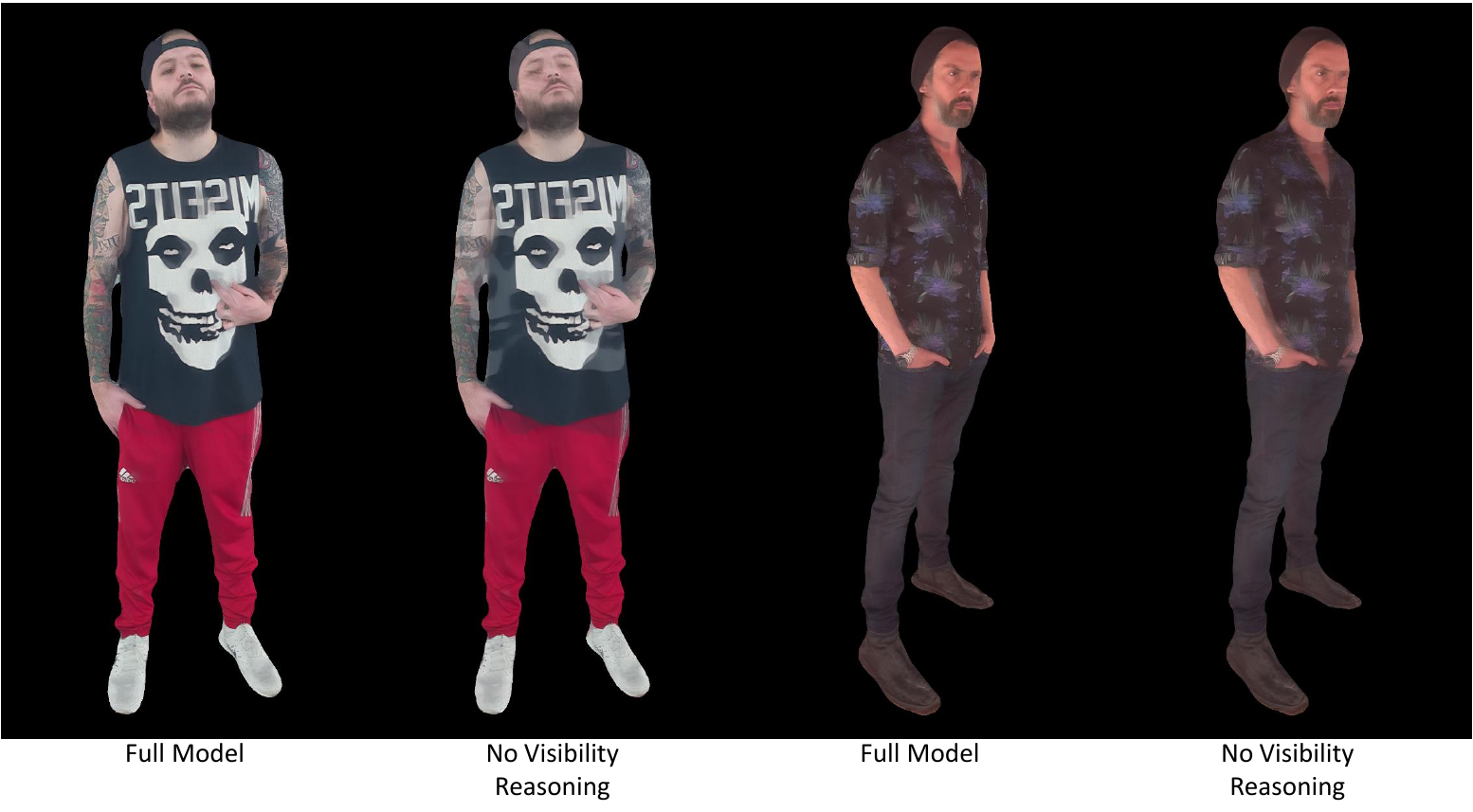}
    \caption{\textbf{Impact of pixel-wise visibility reasoning.} If we warp all the source view pixels to the target view, the rendering results will suffer from severe occlusion problems caused by the wide camera baselines, resulting in blurring and artifacts.}
	\label{fig:vis_ablation}
\end{figure}

\textbf{Impact of the pixel-wise visibility reasoning.} We design a pixel-wise visibility reasoning module to solve the occlusion problems caused by the sparsity of input views. To figure out the strength of this module, we evaluate our method's performance without visibility reasoning. 
Quantitative comparisons on the Twindom dataset are shown in the second and fourth rows of Table~\ref{table:albation}, showing that without visibility reasoning, rendering results will degrade on all three metrics.
Qualitative comparisons are illustrated in Fig.~\ref{fig:vis_ablation}, showing that without pixel-wise visibility reasoning, occluded pixels in the source views will be warped to the target view, resulting in rendering artifacts and blurring.

\begin{figure}
    \centering
  \includegraphics[width=\linewidth]{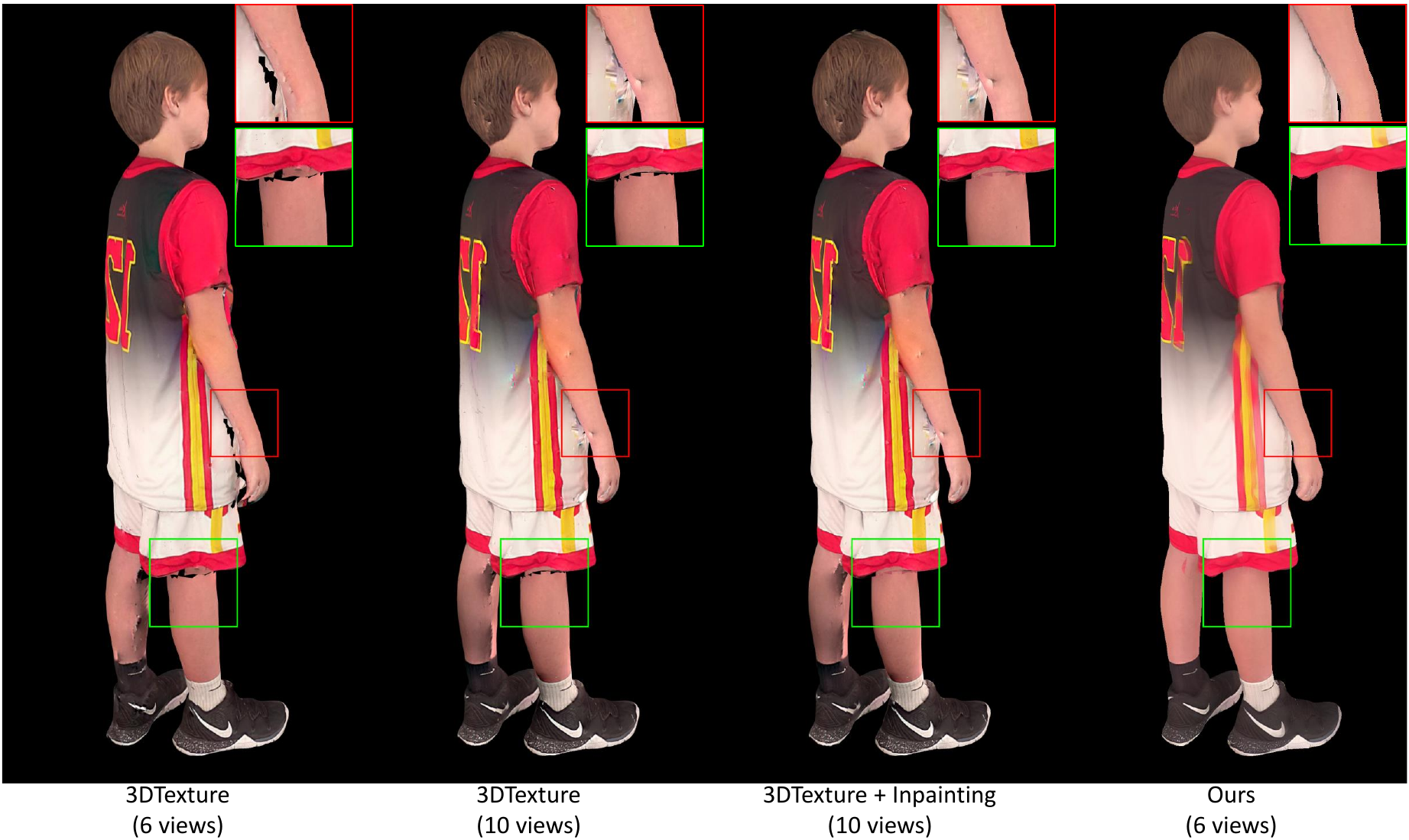}
    \caption{\hl{\textbf{Our rendering network vs. 3DTexture\cite{eccv20143dtexture}.} Compared with 3DTexture that directly "colorizes" the reconstructed mesh, our rendering network is able to render the occluded areas that are invisible in the input views.}}
	\label{fig:abla_3dtexture}
\end{figure}

\begin{figure}[]
    \centering
		\includegraphics[width=\linewidth]{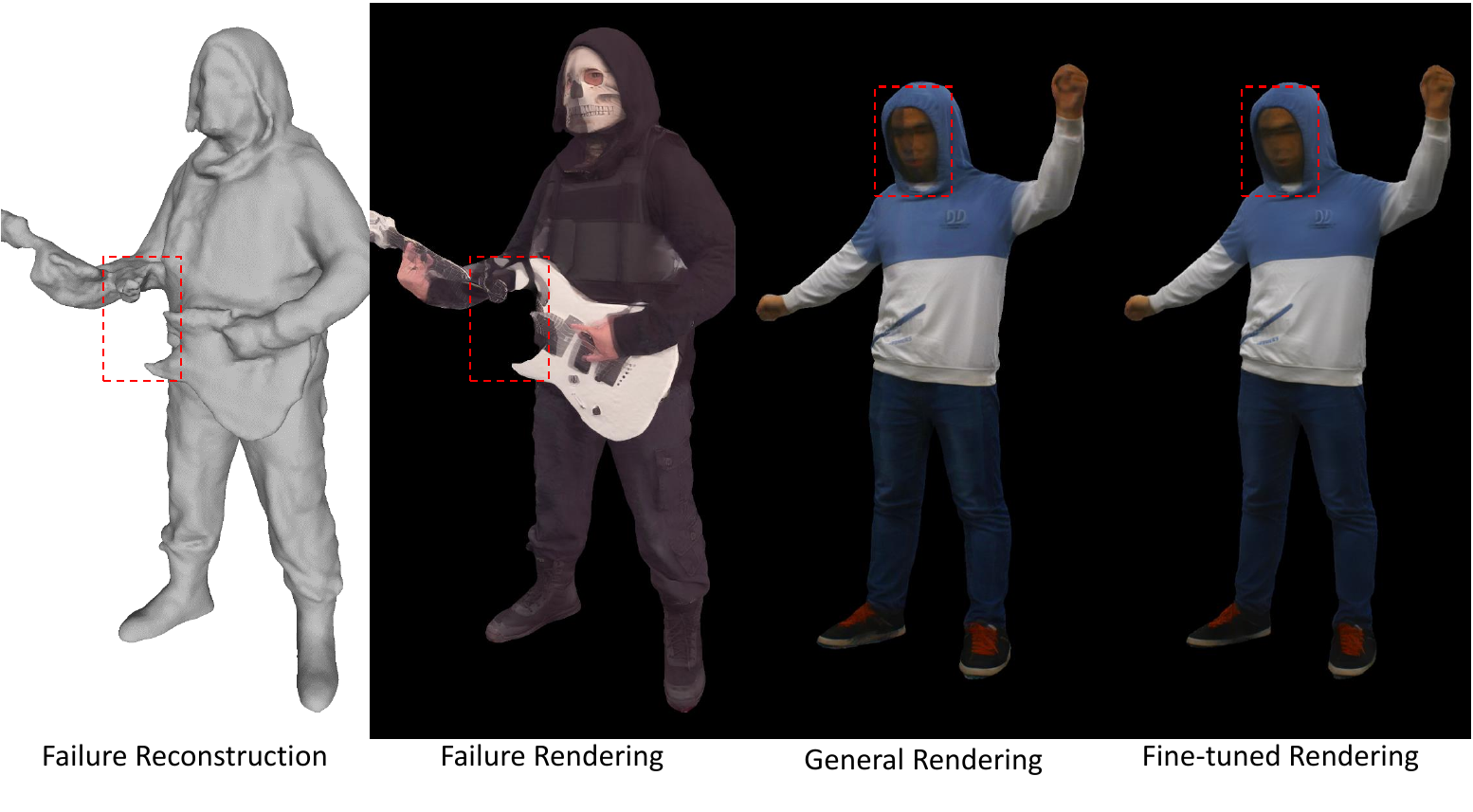}
    \caption{\textbf{Failure cases of our proposed method.
    } The first and second columns show that failure reconstruction results will cause rendering artifacts. The third and fourth columns show that fine-tuning is not an efficient way to improve the rendering quality as we are unable to refine the geometry in the process of fine-tuning. 
    }
	\label{fig:limits}
\end{figure}

\begin{table}[t]
    \caption{\hl{\textbf{Quantitative evaluation of different number of input views.}}}
    \centering
    \small
    \tabcolsep=0.150cm
    \scalebox{0.95}{
        \begin{tabular}{c|ccc|ccc}
        \toprule
        & \multicolumn{3}{c |}{SSIM~$\uparrow$} & 
        \multicolumn{3}{c }{PSNR~$\uparrow$ }\\
        & 6 view & 8 view & 10 view & 6 view & 8 view & 10 view \\
        \hline
        Subject1 & 
         0.876 & 0.884 & \textbf{0.885}      & 25.32 & 25.58 & \textbf{25.63} \\
        Subject2 & 
         0.782 & 0.794 & \textbf{0.796}     & 17.53 & 17.77 & \textbf{17.82} \\
        Subject3  & 
         0.865 & 0.873 & \textbf{0.874}      & 21.75 & 21.97 & \textbf{22.03} \\
        Subject4  &
         0.610 & \textbf{0.626} & 0.625      & 18.53 & 18.70 & \textbf{18.70} \\
        Subject5  & 
         0.794 & \textbf{0.804} & 0.803      & 21.76 & \textbf{21.90} & 21.88 \\
        \hline                                      
        Average & 
         0.785 & 0.796 & \textbf{0.797}      & 20.98 & 21.18 & \textbf{21.21} \\
        \bottomrule
        \end{tabular}
    }
\label{table:abla_num_of_view}
\end{table}

\begin{table}[t]
  \centering
  \caption{\hll{\textbf{Quantitative comparisons with 3DTexture \cite{eccv20143dtexture}.}}}
    \begin{tabular}{c|c|c|c}
    \toprule
          & PSNR~$\uparrow$ & SSIM~$\uparrow$ & LPIPS~$\downarrow$ \\
    \hline
    Ours~(6 views)  & \textbf{20.78} & \textbf{0.856} & \textbf{0.141} \\
    3DTexture~(6~views) & 19.60  & 0.822  & 0.160  \\
    3DTexture + Inpainting~(6~views) & 19.70  & 0.824  & 0.158  \\
    3DTexture~(10~views) & 19.77  & 0.829  & 0.149  \\
    3DTexture + Inpainting~(10~views) & 19.80  & 0.832  & 0.145  \\
    \bottomrule
    \end{tabular}%
  \label{table:abla_texture_mapping}
\end{table}%

\hl{\textbf{Impact of the number of input views.}
The quantitative evaluation on the Twindom dataset with different numbers of input views is illustrated in Table \ref{table:abla_num_of_view}, showing that increasing the number of input views only has slight improvement on the rendering results. Therefore, we can draw a conclusion that our method works well when the input views are highly sparse.}

\hl{\textbf{Rendering network vs. texture mapping.}
To compare with the traditional texture mapping methods and figure out why the rendering network is needed, we use the state-of-the-art texture mapping method 3DTexture\cite{eccv20143dtexture} to ``colorize'' the reconstructed mesh and use OpenGL to render novel-view images. 
\hll{
Qualitative comparisons are illustrated in Fig. \ref{fig:abla_3dtexture}, showing that 3DTexture fails to render the occluded areas that are invisible in the input views, resulting in "black holes" in the invisible areas. 
We also compare an inpainted version for 3DTexture by using an inpainting technique \cite{inpainting} to fill the holes. It can be seen that the hole filling cannot deal with all the artifact problems, while our method is able to produce more reasonable rendering results. Our rendering network can be regarded as a texture prior, which helps to complete these invisible areas.}

Quantitative evaluation on the Twindom dataset is reported in Table \ref{table:abla_texture_mapping}, showing that even with less input views, our rendering network is able to outperform the state-of-the-art texture mapping method.
\hll{
Besides, when dealing with dynamic cases (real-world dataset), as shown in the supplementary video, 3DTexture seems to be sensitive to the reconstruction results. Experiments show that slight flicking of the reconstruction results will lead to incoherence of the rendering results. Moreover, in the areas that suffer from reconstruction errors, the flicking problems will get worse. In contrast, our method is able to produce more stable results on 4D rendering. 
}}

\subsection{Limitations}
As our rendering network needs to use the reconstruction results to perform geometry-guided feature integration, the rendering quality heavily relies on the reconstruction results. If the reconstruction network fails in some parts of the human performer, it will result in rendering artifacts in these parts. The first and second columns of Fig.~\ref{fig:limits} demonstrate that failure reconstruction will cause rendering artifacts in the corresponding parts.

To improve the rendering quality in the parts where the reconstruction network fails, a straightforward way is to fine-tune the rendering network to a specific human performer. However, the geometry representation we used is mesh, and hence the geometry-guided pixel-wise feature integration used in our method is indifferentiable to the reconstructed geometry, making us unable to refine the geometry in the process of fine-tuning. Thus, fine-tuning will cause burring in the parts that suffer from rendering artifacts. The third and fourth columns of Fig.~\ref{fig:limits} show the rendering results that are rendered with the general model and the fine-tuned model, respectively. It can be seen that fine-tuning is unable to solve the rendering artifacts caused by reconstruction failure and will result in blurring in these parts.
\section{Conclusion and Discussion}
In this paper, we introduce HDhuman, a generic method that uses pixel-aligned spatial transformer and geometry-guided pixel-wise feature integration for high-quality human reconstruction and rendering using sparse views. Experiments show that our rendering quality significantly outperforms both the prior generic and specific methods. Our work demonstrates that high-quality human rendering in sparse views is possible without any network fine-tuning, which can serve as an important baseline in the area of human rendering.

In future, a research direction that is worthy to explore is to use a geometry representation that is differentiable in our framework. For example, we could encode the reconstructed mesh to a volume, and then perform differentiable geometry-guided visibility reasoning for each voxel and finally use volume rendering techniques to render novel views. In this way, the geometry will be differentiable, enabling us to refine the geometry in the process of fine-tuning.

\section{Acknowledgments}
This work was supported in part by the National Natural Science Foundation of China (62122058, 62171255, 62171317) and Guoqiang Institute of Tsinghua University (2021GQG0001).

\bibliographystyle{IEEEtran}
\bibliography{paper}

\begin{thebibliography}{10}
\providecommand{\url}[1]{#1}
\csname url@samestyle\endcsname
\providecommand{\newblock}{\relax}
\providecommand{\bibinfo}[2]{#2}
\providecommand{\BIBentrySTDinterwordspacing}{\spaceskip=0pt\relax}
\providecommand{\BIBentryALTinterwordstretchfactor}{4}
\providecommand{\BIBentryALTinterwordspacing}{\spaceskip=\fontdimen2\font plus
\BIBentryALTinterwordstretchfactor\fontdimen3\font minus \fontdimen4\font\relax}
\providecommand{\BIBforeignlanguage}[2]{{%
\expandafter\ifx\csname l@#1\endcsname\relax
\typeout{** WARNING: IEEEtran.bst: No hyphenation pattern has been}%
\typeout{** loaded for the language `#1'. Using the pattern for}%
\typeout{** the default language instead.}%
\else
\language=\csname l@#1\endcsname
\fi
#2}}
\providecommand{\BIBdecl}{\relax}
\BIBdecl

\bibitem{nerf}
B.~Mildenhall, P.~P. Srinivasan, M.~Tancik, J.~T. Barron, R.~Ramamoorthi, and R.~Ng, ``{NeRF}: Representing scenes as neural radiance fields for view synthesis,'' in \emph{Proc. European Conference on Computer Vision}, 2020, pp. 405--421.

\bibitem{neuralbody}
S.~Peng, Y.~Zhang, Y.~Xu, Q.~Wang, Q.~Shuai, H.~Bao, and X.~Zhou, ``Neural body: Implicit neural representations with structured latent codes for novel view synthesis of dynamic humans,'' in \emph{Proc. IEEE/CVF Conference on Computer Vision and Pattern Recognition}, 2021.

\bibitem{smpl}
M.~Loper, N.~Mahmood, J.~Romero, G.~Pons-Moll, and M.~J. Black, ``{SMPL}: A skinned multi-person linear model,'' \emph{ACM Transactions on Graphics}, vol.~34, no.~6, pp. 1--16, 2015.

\bibitem{pifu}
S.~Saito, , Z.~Huang, R.~Natsume, S.~Morishima, A.~Kanazawa, and H.~Li, ``{PIFu}: Pixel-aligned implicit function for high-resolution clothed human digitization,'' in \emph{Proc. IEEE/CVF International Conference on Computer Vision}, 2019.

\bibitem{dynamicTexture}
L.~Liu, W.~Xu, M.~Habermann, M.~Zollhöfer, F.~Bernard, H.~Kim, W.~Wang, and C.~Theobalt, ``Learning dynamic textures for neural rendering of human actors,'' \emph{IEEE Transactions on Visualization and Computer Graphics}, vol.~27, no.~10, pp. 4009--4022, 2021.

\bibitem{neuraltexture}
J.~Thies, M.~Zollh{\"o}fer, and M.~Nie{\ss}ner, ``Deferred neural rendering: Image synthesis using neural textures,'' \emph{ACM Transactions on Graphics}, vol.~38, no.~4, pp. 1--12, 2019.

\bibitem{continuousdepth}
Y.~Liu, X.~Cao, Q.~Dai, and W.~Xu, ``Continuous depth estimation for multi-view stereo,'' in \emph{Proc. IEEE/CVF Conference on Computer Vision and Pattern Recognition}, 2009, pp. 2121--2128.

\bibitem{highquality}
A.~Collet, M.~Chuang, P.~Sweeney, D.~Gillett, D.~Evseev, D.~Calabrese, H.~Hoppe, A.~Kirk, and S.~Sullivan, ``High-quality streamable free-viewpoint video,'' \emph{ACM Transactions on Graphics}, vol.~34, no.~4, pp. 1--13, 2015.

\bibitem{fusion4d}
M.~Dou, S.~Khamis, Y.~Degtyarev, P.~Davidson, S.~R. Fanello, A.~Kowdle, S.~O. Escolano, C.~Rhemann, D.~Kim, J.~Taylor \emph{et~al.}, ``{Fusion4D}: Real-time performance capture of challenging scenes,'' \emph{ACM Transactions on Graphics}, vol.~35, no.~4, pp. 1--13, 2016.

\bibitem{function4d}
T.~Yu, Z.~Zheng, K.~Guo, P.~Liu, Q.~Dai, and Y.~Liu, ``{Function4D}: Real-time human volumetric capture from very sparse consumer {RGBD} sensors,'' in \emph{Proc. IEEE/CVF Conference on Computer Vision and Pattern Recognition}, 2021, pp. 5746--5756.

\bibitem{realtime}
K.~Guo, F.~Xu, T.~Yu, X.~Liu, Q.~Dai, and Y.~Liu, ``Real-time geometry, albedo, and motion reconstruction using a single {RGB}-{D} camera,'' \emph{ACM Transactions on Graphics}, vol.~36, no.~4, p.~1, 2017.

\bibitem{fof}
Q.~Feng, Y.~Liu, Y.-K. Lai, J.~Yang, and K.~Li, ``{FOF}: Learning fourier occupancy field for monocular real-time human reconstruction,'' in \emph{NeurIPS}, 2022.

\bibitem{image}
K.~Li, H.~Wen, Q.~Feng, Y.~Zhang, X.~Li, J.~Huang, C.~Yuan, Y.-K. Lai, and Y.~Liu, ``Image-guided human reconstruction via multi-scale graph transformation networks,'' \emph{IEEE Transactions on Image Processing}, vol.~30, pp. 5239--5251, 2021.

\bibitem{learning}
X.~Li, J.~Huang, J.~Zhang, X.~Sun, H.~Xuan, Y.-K. Lai, Y.~Xie, J.~Yang, and K.~Li, ``Learning to infer inner-body under clothing from monocular video,'' \emph{IEEE Transactions on Visualization and Computer Graphics}, 2022.

\bibitem{pifuhd}
S.~Saito, T.~Simon, J.~Saragih, and H.~Joo, ``{PIFuHD}: Multi-level pixel-aligned implicit function for high-resolution {3D} human digitization,'' in \emph{Proc. IEEE/CVF Conference on Computer Vision and Pattern Recognition}, 2020, pp. 84--93.

\bibitem{livecap}
M.~Habermann, W.~Xu, M.~Zollhoefer, G.~Pons-Moll, and C.~Theobalt, ``{LiveCap}: Real-time human performance capture from monocular video,'' \emph{ACM Transactions on Graphics}, vol.~38, no.~2, pp. 1--17, 2019.

\bibitem{monoperfcap}
W.~Xu, A.~Chatterjee, M.~Zollh{\"o}fer, H.~Rhodin, D.~Mehta, H.-P. Seidel, and C.~Theobalt, ``{MonoPerfCap}: Human performance capture from monocular video,'' \emph{ACM Transactions on Graphics}, vol.~37, no.~2, pp. 1--15, 2018.

\bibitem{smplicit}
E.~Corona, A.~Pumarola, G.~Alenya, G.~Pons-Moll, and F.~Moreno-Noguer, ``{SMPLicit}: Topology-aware generative model for clothed people,'' in \emph{Proc. IEEE/CVF Conference on Computer Vision and Pattern Recognition}, 2021, pp. 11\,875--11\,885.

\bibitem{monoclothcap}
D.~Xiang, F.~Prada, C.~Wu, and J.~Hodgins, ``{MonoClothCap}: Towards temporally coherent clothing capture from monocular video,'' in \emph{Proc. International Conference on 3D Vision}, 2020, pp. 322--332.

\bibitem{zins}
P.~Zins, Y.~Xu, E.~Boyer, S.~Wuhrer, and T.~Tung, ``Data-driven {3D} reconstruction of dressed humans from sparse views,'' in \emph{Proc. International Conference on 3D Vision}.\hskip 1em plus 0.5em minus 0.4em\relax IEEE, 2021, pp. 494--504.

\bibitem{deepmulticap}
Y.~Zheng, R.~Shao, Y.~Zhang, T.~Yu, Z.~Zheng, Q.~Dai, and Y.~Liu, ``{DeepMultiCap}: Performance capture of multiple characters using sparse multiview cameras,'' \emph{arXiv preprint arXiv:2105.00261}, 2021.

\bibitem{deepblending}
P.~Hedman, J.~Philip, T.~Price, J.-M. Frahm, G.~Drettakis, and G.~Brostow, ``Deep blending for free-viewpoint image-based rendering,'' \emph{ACM Transactions on Graphics}, vol.~37, no.~6, pp. 1--15, 2018.

\bibitem{depthsynthesis}
G.~Chaurasia, S.~Duchene, O.~Sorkine-Hornung, and G.~Drettakis, ``Depth synthesis and local warps for plausible image-based navigation,'' \emph{ACM Transactions on Graphics}, vol.~32, no.~3, pp. 1--12, 2013.

\bibitem{ignor}
J.~Thies, M.~Zollh{\"o}fer, C.~Theobalt, M.~Stamminger, and M.~Nie{\ss}ner, ``{IGNOR}: Image-guided neural object rendering,'' \emph{arXiv preprint arXiv:1811.10720}, 2018.

\bibitem{pointmvs}
Y.~Liu, Q.~Dai, and W.~Xu, ``A point-cloud-based multiview stereo algorithm for free-viewpoint video,'' \emph{IEEE Transactions on Visualization and Computer Graphics}, vol.~16, no.~3, pp. 407--418, 2009.

\bibitem{fvs}
G.~Riegler and V.~Koltun, ``Free view synthesis,'' in \emph{Proc. European Conference on Computer Vision}, 2020.

\bibitem{neuralhumanfvv}
X.~Suo, Y.~Jiang, P.~Lin, Y.~Zhang, M.~Wu, K.~Guo, and L.~Xu, ``Neuralhumanfvv: Real-time neural volumetric human performance rendering using {RGB} cameras,'' in \emph{Proc. IEEE/CVF Conference on Computer Vision and Pattern Recognition}, 2021, pp. 6226--6237.

\bibitem{stereomag}
T.~Zhou, R.~Tucker, J.~Flynn, G.~Fyffe, and N.~Snavely, ``Stereo magnification: Learning view synthesis using multiplane images,'' \emph{arXiv preprint arXiv:1805.09817}, 2018.

\bibitem{soft3d}
E.~Penner and L.~Zhang, ``Soft {3D} reconstruction for view synthesis,'' \emph{ACM Transactions on Graphics}, vol.~36, no.~6, pp. 1--11, 2017.

\bibitem{llff}
B.~Mildenhall, P.~P. Srinivasan, R.~Ortiz-Cayon, N.~K. Kalantari, R.~Ramamoorthi, R.~Ng, and A.~Kar, ``Local light field fusion: Practical view synthesis with prescriptive sampling guidelines,'' \emph{ACM Transactions on Graphics}, vol.~38, no.~4, pp. 1--14, 2019.

\bibitem{deepview}
J.~Flynn, M.~Broxton, P.~Debevec, M.~DuVall, G.~Fyffe, R.~Overbeck, N.~Snavely, and R.~Tucker, ``{DeepView}: View synthesis with learned gradient descent,'' in \emph{Proc. IEEE/CVF Conference on Computer Vision and Pattern Recognition}, 2019, pp. 2367--2376.

\bibitem{nsvf}
L.~Liu, J.~Gu, K.~Zaw~Lin, T.-S. Chua, and C.~Theobalt, ``Neural sparse voxel fields,'' in \emph{Proc. International Conference on Neural Information Processing Systems}, 2020.

\bibitem{StructLocalNerf}
Z.~Zheng, H.~Huang, T.~Yu, H.~Zhang, Y.~Guo, and Y.~Liu, ``Structured local radiance fields for human avatar modeling,'' in \emph{Proc. IEEE/CVF Conference on Computer Vision and Pattern Recognition}, June 2022, pp. 15\,893--15\,903.

\bibitem{HumanNerf_Zhao}
F.~Zhao, W.~Yang, J.~Zhang, P.~Lin, Y.~Zhang, J.~Yu, and L.~Xu, ``{HumanNeRF}: Efficiently generated human radiance field from sparse inputs,'' in \emph{Proc. IEEE/CVF Conference on Computer Vision and Pattern Recognition}, 2022, pp. 7743--7753.

\bibitem{SurfaceAlignNerf}
T.~Xu, Y.~Fujita, and E.~Matsumoto, ``Surface-aligned neural radiance fields for controllable 3d human synthesis,'' in \emph{Proc. IEEE/CVF Conference on Computer Vision and Pattern Recognition}, June 2022, pp. 15\,883--15\,892.

\bibitem{AnimatableNerf}
S.~Peng, J.~Dong, Q.~Wang, S.~Zhang, Q.~Shuai, X.~Zhou, and H.~Bao, ``Animatable neural radiance fields for modeling dynamic human bodies,'' in \emph{Proc. IEEE/CVF International Conference on Computer Vision}, 2021, pp. 14\,314--14\,323.

\bibitem{EasyMointeracCap}
Q.~Shuai, C.~Geng, Q.~Fang, S.~Peng, W.~Shen, X.~Zhou, and H.~Bao, ``Novel view synthesis of human interactions from sparse multi-view videos,'' in \emph{Proc. ACM SIGGRAIPH}, 2022, p.~10.

\bibitem{HumanNerf_Weng}
C.-Y. Weng, B.~Curless, P.~P. Srinivasan, J.~T. Barron, and I.~Kemelmacher-Shlizerman, ``{HumanNeRF}: Free-viewpoint rendering of moving people from monocular video,'' in \emph{Proc. IEEE/CVF Conference on Computer Vision and Pattern Recognition}, 2022, pp. 16\,210--16\,220.

\bibitem{dnerf}
A.~Pumarola, E.~Corona, G.~Pons-Moll, and F.~Moreno-Noguer, ``D-nerf: Neural radiance fields for dynamic scenes,'' in \emph{Proceedings of the IEEE/CVF Conference on Computer Vision and Pattern Recognition}, 2021, pp. 10\,318--10\,327.

\bibitem{marchingcube}
W.~E. Lorensen and H.~E. Cline, ``Marching cubes: A high resolution {3D} surface construction algorithm,'' \emph{Proc. ACM SIGGRAIPH}, vol.~21, no.~4, pp. 163--169, 1987.

\bibitem{openpose}
Z.~Cao, G.~Hidalgo, T.~Simon, S.-E. Wei, and Y.~Sheikh, ``{OpenPose}: realtime multi-person {2D} pose estimation using part affinity fields,'' \emph{IEEE Transactions on Pattern Analysis and Machine Intelligence}, vol.~43, no.~1, pp. 172--186, 2019.

\bibitem{SVS}
G.~Riegler and V.~Koltun, ``Stable view synthesis,'' in \emph{Proc. IEEE/CVF Conference on Computer Vision and Pattern Recognition}, 2021, pp. 12\,216--12\,225.

\bibitem{he2016deep}
K.~He, X.~Zhang, S.~Ren, and J.~Sun, ``Deep residual learning for image recognition,'' in \emph{Proc. IEEE/CVF Conference on Computer Vision and Pattern Recognition}, 2016, pp. 770--778.

\bibitem{twindom}
\BIBentryALTinterwordspacing
 [Online]. Available: \url{https://web.twindom.com/}
\BIBentrySTDinterwordspacing

\bibitem{adam}
D.~P. Kingma and J.~Ba, ``{Adam}: A method for stochastic optimization,'' \emph{arXiv preprint arXiv:1412.6980}, 2014.

\bibitem{eccv20143dtexture}
M.~Waechter, N.~Moehrle, and M.~Goesele, ``{Let There Be Color!} large-scale texturing of {3D} reconstructions,'' in \emph{Proc. European Conference on Computer Vision}, 2014, pp. 836--850.

\bibitem{inpainting}
P.~P\'{e}rez, M.~Gangnet, and A.~Blake, ``Poisson image editing,'' in \emph{Proc. ACM SIGGRAPH Asia}, 2003, p. 313–318.

\end{thebibliography}


\end{document}